\documentclass[10pt,compsoc]{IEEEtran}
\usepackage{subfigure}
\usepackage{graphicx}
\usepackage{array}
\usepackage{graphicx}
\usepackage{epsfig}
\usepackage{amsmath}
\usepackage{multirow}
\usepackage{ragged2e}
\usepackage{subfig}
\usepackage{float}
\usepackage{algorithm,algpseudocode}

\renewcommand{\paragraph}[1]{{\smallskip \vspace{1pt} \noindent \bf #1}}

\begin{document}
\title{Scalable Nearest Neighbor Search \\ based on kNN Graph}

\author{Wan-Lei~Zhao*,
        ~Jie~Yang,
        ~Cheng-Hao Deng
\IEEEcompsocitemizethanks{\IEEEcompsocthanksitem Fujian Key Laboratory of Sensing and Computing for Smart City, and the School of Information Science and Engineering, Xiamen University, Xiamen, 361005, P. R. China.\protect\\
Wan-Lei Zhao is the correspoding author. E-mail: wlzhao@xmu.edu.cn. 
}}

\IEEEtitleabstractindextext{
\begin{justify}
\begin{abstract}
Nearest neighbor search is known as a challenging issue that has been studied for several decades. Recently, this issue becomes more and more imminent in viewing that the big data problem arises from various fields. In this paper, a scalable solution based on hill-climbing strategy with the support of k-nearest neighbor graph (kNN) is presented. Two major issues have been considered in the paper. Firstly, an efficient kNN graph construction method based on two means tree is presented. For the nearest neighbor search, an enhanced hill-climbing procedure is proposed, which sees considerable performance boost over original procedure. Furthermore, with the support of inverted indexing derived from residue vector quantization, our method achieves close to 100\% recall with high speed efficiency in two state-of-the-art evaluation benchmarks. In addition, a comparative study on both the compressional and traditional nearest neighbor search methods is presented. We show that our method achieves the best trade-off between search quality, efficiency and memory complexity.
\end{abstract}
\end{justify}
}

\maketitle

%\IEEEpeerreviewmaketitle
\section{Introduction}
Nearest neighbor search (NNS) generally arises from a wide range of subjects, such as database, machine learning, computer vision, and information retrieval. Due to the fundamental role it plays, it has been studied in many computer fields for several decades. The nearest neighbor search problem can be simply defined as follows. Given a query vector ($q \in R^D$), and \textit{n} candidates that are under the same dimensionality. It is required to return sample(s) for the query that are spatially closest to it according to certain metric (usually $\textit{l}_1$-distance or $\textit{l}_2$-distance).

Traditionally, this issue has been addressed by various space partitioning strategies. However, these methods are hardly scalable to high dimensional (e.g., $D > 20$), large-scale and dense vector space. In such case, most of the traditional methods such as k-d tree~\cite{kdtree75}, R-tree~\cite{rtree84} and locality sensitive hashing (LSH)~\cite{lsh04} are unable to return decent results. 

Recently, there are two major trends in the literature. In one direction, the nearest neighbor search is conducted based on k-nearest neighbor graph (kNN graph)~\cite{icai11:kiana,mm12:jdwang,efanna16}, in which the kNN graph is constructed offline. Alternatively, NNS is addressed based on vector quantization~\cite{JDS11,cvpr14:artem,icml14:tzhang}. The primary goal of this way is to compress the reference set by vector quantization. Such that it is possible to load the whole reference set (after compression) into the memory in the case that the reference set is extremely large. 

There are two key issues to be addressed in kNN graph based methods. The primary issue is about the efficient construction of the kNN graph, given the original problem is at the complexity of $O(D{\cdot}n^2)$. Another issue is how to perform the NN search given kNN graph is supplied. For the first issue, it has been addressed by the scheme ``divide-and-conquer''~\cite{acmcm80:bentley, cvpr12:jingwang} and nearest neighbor descent~\cite{weidong}. The combination of these two schemes is also seen from recent work~\cite{efanna16}. Thanks to the above schemes, the complexity of constructing kNN graph has been reduced to around $O(D{\cdot}n{\cdot}log(n))$. With the support of kNN graph, nearest neighbor search is conducted by hill-climbing strategy~\cite{icai11:kiana}. However, this strategy could be easily trapped in local optima since the seed samples are randomly selected. Recently, this issue has been addressed by search perturbation~\cite{mm12:jdwang}, in which an intervention scheme is introduced as the search procedure is trapped in a local optima. As indicated in recent work~\cite{efanna16}, this solution is sub-optimal. Alternatively, with the support of k-d trees to index the whole reference set, the hill climbing search is directed to start from the neighborhood of potential nearest neighbors~\cite{efanna16}. However, in order to achieve high performance, multiple k-d trees are required, which causes considerable memory overhead.

Considering the big memory cost with the traditional methods, NNS is addressed based on vector quantization. The representative methods are product quantizer (PQ)~\cite{JDS11}, additive quantizer (AQ)~\cite{cvpr14:artem}, composite quantizer (CQ)~\cite{icml14:tzhang} and stacked quantizer (SQ)~\cite{sq14}. Essentially in all these methods, vectors in the reference set are compressed via vector quantization. The advantages that the quantization methods bring are two folds. On one hand, the candidate vectors have been compressed (typically the memory consumption is one order of magnitude lower than the size of reference set), which makes it possible to scale-up the nearest neighbor search to very large range. On the other hand, the distance computation between query and all the candidates becomes very efficient when it is approximated by the distances between query and vocabulary words. However, due to the approximation in both vector representation and distance calculation, high recall is undesirable with the compressional methods.

In our paper, similar as~\cite{icai11:kiana}~\cite{mm12:jdwang}~\cite{efanna16}, NNS is basically addressed via hill-climbing strategy based on kNN graph. Two major issues are considered. Firstly, an efficient and scalable kNN graph construction method is presented. In addition, different from existing strategies, the hill-climbing search is directed by multi-layer residue vector quantization (RVQ)~\cite{rq96:barnes}. The residue quantization forms a coarse-to-fine partition over the whole vector space. For this reason, it is able to direct the hill-climbing search to start most likely from within the neighborhood of the true nearest neighbor. Since the inverted indexing derived from residue quantization only keeps the IDs of the reference vectors, in comparison to method in~\cite{efanna16}, ignorable extra memory is required. As a result, the major memory cost of our method is to hold the reference set. Furthermore, with the support of the multi-layer quantization, sparse indexing over the big reference set is achievable with a series of small vocabularies, which makes the search procedure very efficient and easily scale-up to billion level dataset.

The remaining of this paper is organized as follows. In Section~\ref{sec:rela}, a brief review about the literature of NNS is presented. Section~\ref{sec:nns} presents an efficient algorithm for kNN graph construction. Based on the constructed kNN graph, an enhanced hill-climbing procedure for nearest neighbor search is therefore presented. Section~\ref{sec:invt} shows how the inverted indexing structure is built based on residue vector quantization, which will boost the performance of hill-climbing procedure. The experimental study over the enhanced hill-climbing nearest neighbor search is presented in Section~\ref{sec:exp}. 

\section{Related Works}
\label{sec:rela}
The early study about the issue of NNS could be traced back to 1970s, when the need of NNS search over file system arises. In those days, the data to be processed are in very low dimension, typically 1D. This problem is well-addressed by B-Tree~\cite{btree:commer79} and its variants $B^+$-tree~\cite{btree:commer79}, based on which the NNS complexity could be as low as $O(log(n))$. However, B-tree is not naturally extensible to more than 1D case. More sophisticated indexing structures were designed to handle NNS in multi-dimensional data. Representative structures are k-d-tree~\cite{kdtree75}, R-tree~\cite{rtree84} and R*-tree~\cite{BKSB90}. For k-d tree, pivot vector is selected each time to split the dataset evenly into two. By applying this bisecting repeatedly, the hyper-space is partitioned into embedded hierarchical sub-spaces. The NNS is performed by tranversing over one or several branches to probe the nearest neighbor. The space partitioning aims to restrict the search taking place within minimum number of sub-spaces. However, unlike B-tree in 1D case, the partition scheme does not exclude the possibility that nearest neighbor resides outside of these candidate sub-spaces. Therefore, extensive probing over the large number of branches in k-d tree becomes inevitable. For this reason, NNS with k-d tree could be very slow. Similar comments apply to R-tree and its variants despite that a more sophisticated design on the partition strategy is proposed in these tree structures. Recent indexing structure FLANN~\cite{pami14:flann} partitions the space with hierarchical k-means and multiple k-d trees. Although efficient, sub-optimal results are achieved.

For all the aforementioned tree partitioning methods, another major disadvantage lies in their heavy demand in memory. On one hand, in order to support fast comparison, all the candidate vectors are loaded into the memory. On the other hand, the tree nodes that are used for indexing also take up considerable amount of extra memory. Overall, the memory consumption is usually several times bigger than the size of candidate set.

Apart from tree partitioning method, several attempts have been made to apply locality sensitive hashing (LSH)~\cite{lsh04} on NNS. In general, there are two steps involved in the search stage. Namely, step 1. collects the candidates that share the same or similar hash keys as the query; step 2. performs exhaustive comparison between the query and all these selected candidates to find out the nearest neighbor. Similar as FLANN, LSH is good for the application that only requires approximate nearest neighbor. Moreover, in order to support the fast comparison in step 2, the whole reference set are required to be loaded in the memory, which leads to a lot of memory consumption. 

In recent years, the problem of NNS is addressed by vector quantization~\cite{tit06:gray06}. The proposal of applying product quantizer~\cite{JDS11} on NNS opens up new way to solve this decades old problem. For all the quantization based methods~\cite{ivfrvq10, JDS11,artem16,icml14:tzhang, sq14}, they share two things in common. Firstly, the candidate vectors are all compressed via vector quantization. This makes it easier than previous methods to hold the whole dataset in the memory. Secondly, NNS is conducted between the query and the compressed candidate vectors. The distance between query and candidates is approximated by the distance between query and vocabulary words that are used for quantization. Due to the heavy compression on the reference vectors, high recall on the top is hardly desirable with these methods.

Recently, the hill-climbing strategy has been also explored~\cite{icai11:kiana,mm12:jdwang,efanna16}. Basically, no sophisticated space partitioning is required in the original idea~\cite{icai11:kiana}. The NNS starts from a group of random seeds (random location in the vector space). The iterative procedure tranverses over the kNN graph by depth-first search. Guided by kNN graph, the search procedure ascents closer to the true nearest neighbor in each round. It takes only few number of rounds to converge. Since the procedure starts from random position, it is likely to be trapped in local optima. To alleviate this issue, an intervention scheme is introduced in~\cite{mm12:jdwang}. However, this method turns out to be sub-optimal. Recently, this issue is addressed by indexing the reference set by multiple k-d trees, which is similar as FLANN~\cite{pami14:flann}. When one query comes, the search tranverses over these k-d trees. Potential candidates are collected as the seeds for hill-climbing search. Such that the search process could be faster and the possibility that is trapped in a local optima is lower. Unfortunately, such kind of indexing causes nearly one times of memory overhead to load the k-d trees, which makes it inscalable to large-scale search task.

Hill-climbing strategy is adopted in our search scheme due to its simplicity and efficiency. Different from~\cite{mm12:jdwang,efanna16}, the search procedure is directed by residue vector quantization (RVQ)~\cite{ivfrvq10}. Firstly, the candidate vectors are quantized by RVQ. However, different from compressional methods, the quantized codes are used only to build inverted indexing. Candidate vectors sharing the same RVQ codes (indexing key) are organized into one inverted list. In comparison to~\cite{efanna16}, only 4 bytes are required to keep the index for one vector. When a query comes, the distance between the query and the RVQ codes are calculated, the candidates reside in the lists of top-ranked codes are taken as the seeds for hill-climbing procedure. The inverted indexing built from RVQ codes plays the similar role as k-d trees in~\cite{efanna16}, while requiring very small amount of memory.

\section{NNS based on k-NN Graph Construction}
\label{sec:nns}
\subsection{k-NN Graph Construction with Two Means Clustering}
\label{sec:knn}
In this section, a novel kNN graph construction method based on two means tree~\cite{2mtree} is presented. Basically, we follow the strategy of divide-and-conquer~\cite{acmcm80:bentley, cvpr12:jingwang, efanna16}. While different from~\cite{acmcm80:bentley, cvpr12:jingwang, efanna16}, the division over the vector space is undertaken by two means clustering in our paper. The motivation behind this algorithm is based on the observation that samples in one cluster are most likely neighbors. The general procedure that constructs the kNN graph is in two steps. Firstly, the reference set is divided into small clusters, where the size of each cluster is no bigger than a threshold (i.e., 50). This could be efficiently fulfilled by bisecting the reference set repeatedly with two means clustering. In the second step, an exhaustive distance comparison is conducted within each cluster. The kNN list for each cluster member is therefore built or updated. This general procedure could be repeated for several times to refine the kNN list of each sample.

\begin{algorithm}{kNNG construction}
  \begin{algorithmic}[1]
    \State \textbf{Input}: $X_{n{\times}d}$: reference set, k: scale of k-NN Graph
    \State \textbf{Output}: kNN Graph $G_{n{\times}k}$
    \State Initialize $G_{n{\times}k}$; t $\leftarrow$ 0;
    \For {t $<$ $I_0$}
    	\State Apply two means clustering on $X_{n{\times}d}$
    	\State Collect clusters set $S$
	    \For {each $S_m \in S$}
		    \For {each $<i, j>$} //($i, j \in S_m$ \& $i {\neq}j$)
  			    \If { $<i, j>$ is NOT visited}
				    \State Update $G[i]$ and $G[j]$ with $d(x_i, x_j)$;
				 \EndIf
		    \EndFor
		\EndFor
		\State t $\leftarrow$ t + 1
    \EndFor
  \end{algorithmic}
  \label{alg:knng}
\end{algorithm}

As shown in Alg.~\ref{alg:knng}, the kNN graph construction mainly relies on two means clustering, which forms a partition on the input reference set in each round. Comparing with k-d tree, two means produces ball shape clusters instead of cubic shape partition. In order to achieve high quality, the clustering is driven by objective function of boost k-means~\cite{boostkmeans}, which always leads to higher clustering quality. With the clustering result of each round, the brute-force comparison is conducted within each cluster. Due to the small scale of each cluster, this process could be very efficient. The kNN graph $G$ is refined as new and closer neighbors join in. Notice that the result of two means clustering is different from one round to another. We find it is sufficient to set $I_0$ to $10$ for million level dataset. The overall complexity of this procedure is $O({I_0}{\cdot}n{\cdot}log(n){\cdot}D)$, where $n{\cdot}log(n){\cdot}D$ is actually the complexity of two means clustering.

\subsection{Enhanced Hill-climbing Procedure for NNS}
Once the k-NN graph is ready, it is quite natural to adopt the hill-climbing scheme~\cite{icai11:kiana} to perform the NNS. However, we find that the procedure proposed in~\cite{icai11:kiana} is easily trapped in local optima and involves redundant visits to unlikely NN candidates. For this reason, the procedure is revised slightly in the paper, which is given in Alg.~\ref{alg:egnns}.

\begin{algorithm}{Ehanced Hill-Climbing NNS}  
  \begin{algorithmic}[1]
    \State \textbf{Input}: q: query, G: k-NN Graph,\\ S: seeds, $R{\leftarrow} \phi$: NN rank-list
    \State \textbf{Output}:  R
    \State $R{\leftarrow}R{\cup}$ \{$<$s, d(q, s)$>$\}, $s \in S$; $t{\leftarrow}0$;
    \While {t $< t_0$}
	    \State T $\leftarrow$ $\phi$;
    	\For {each $r_i \in$ top-k of R}
       		\For {each $n_j \in G[r_i]$}
       			\If { $n_j$ is NOT visited}
		    		\State $T{\leftarrow}T{\cup}$ \{$<n_j, d(q, n_j)>$\};
		    	\EndIf
			\EndFor
    	\EndFor
    	\State $R{\leftarrow}R{\cup}T$;
        \State $t{\leftarrow}t+1$;
    \EndWhile
  \end{algorithmic}
 \label{alg:egnns}
\end{algorithm}
In Alg.~\ref{alg:egnns}, seeds could be randomly selected as~\cite{icai11:kiana} does or supplied by inverted indexing, which will be introduced in Section~\ref{sec:invt}. As will be seen in the experiments, seeds supplied by inverted indexing lead to significant performance boost. 

As shown in Alg.~\ref{alg:egnns}, the major modification on the original algorithm is on the way of kNN expansion. As seen from Line 8-14, the NN expansion is undertaken for each top-k sample in one iteration. In contrast, algorithm given in~\cite{icai11:kiana} expands kNN only for the top-ranked sample in one iteration, which does not allow all the seeds to hold equal chance to climb the hill. We find that the original procedure is more likely trapped in local optima. As will be seen in the experiment, more number of iterations is needed to reach to the same level of search quality as our enhanced version.

\section{Fast Nearest Neighbor Search}
\label{sec:invt}
\subsection{Review on Residue Quantization}
The idea of approximating vectors by the composition of several vectors could be traced back to the design of ``two-stage residue vector quantization''~\cite{vq84:gray}. In this scheme, an input vector is encoded by a quantizer, and its residue (quantization error). The residue vector is sequentially encoded by another quantizer. This scheme has been extended to several orders (stages)~\cite{rq96:barnes}, in which the residue vector that is left from the prior quantization stage is quantized recursively. Given vector $v \in R^D$, and a series vocabularies $\{V_1, \cdots, V_k\}$, vector $v$ is approximated by a composition of words from these vocabularies. In particular, one word is selected from one stage vocabulary. Given this recursive quantization is repeated to \textit{k}-th stage, vector $v$ is approximated as follows.
\begin{equation}
 v \approx w^{(1)}_{i_1}+\cdots+w^{(m)}_{i_m}+\cdots+w^{(k)}_{i_k}
\label{eqn:rq}
\end{equation}
In Eqn.~\ref{eqn:rq}, $w^{(m)}_{i_m} \in V_m$ in stage \textit{m} are trained on the residues $r=v-\sum_{j=1}^{(m-1)}{w^{(j)}_{i_j}}$ collected from $m-1$ stage. Finally, vector $v$ after RVQ is encoded as ${c_1}{\cdots}{c_i}{\cdots}c_m$, where $c_i$ is the word ID from vocabulary $V_i$. When only one stage is employed, residue quantization is nothing more than vector quantization, which has been well-known as Bag-of-visual word model~\cite{SiZ03}. 

As discussed in~\cite{ivfrvq10}, multiple-stage RVQ forms a coarse-to-fine partition over the vector space, which plays similar role as hierarchical k-means (HKM) in FLANN~\cite{pami14:flann}. While different from HKM, the space complexity of RVQ is very small given there is no need to keep the big amount of high-dimensional tree nodes. In addition, RVQ is able to produce a big amount of distinctive codes with small vocabularies as multiple stages are employed. For instance, given $m=4, |V_{i=1{\cdots}4}|=256$, the number of distinctive codes is as much as $2^{32}$.

\subsection{Inverted Indexing with RVQ and Cascaded Pruning}
\label{sec:invt1}
In RVQ encoding, the energy of a vector is mainly kept in the codes of lower order. Similar as~\cite{cvpr12:artem, ivfrvq10}, in our design, the codes from multiple stage RVQ are combined as the indexing key, which indicates the rough location of a vector in the space. Such that all the vectors that reside close to this location share the same indexing key. The more orders of codes are incorporated, the more precise this location and the larger of the inverted indexing space. Consequently, the encoded vectors are sparsely distributed in the inverted lists. 

When a query comes, the search process calculates the distance between query and indexing keys. On the first hand, the indexing keys are decomposed back into several RVQ codes. The asymmetric distance between the query and these codes are therefore calculated and sorted. Given query \textit{q} and the indexing key $I$ formed by the first two order RVQ codes (i.e., $I={c_1c_2}$), the distance between \textit{q} and key $I$ is given as
\begin{align}
	d(q, I) =& (q - w^{(1)}_{c_1}-w^{(2)}_{c_2})^2 \nonumber \\
	          =& q{\cdot}q^t - 2{\cdot}(w^{(1)}_{c_1} + w^{(2)}_{c_2}){\cdot}q^t+(w^{(1)}_{c_1}+w^{(2)}_{c_2})^2,
\label{eqn:q2key}
\end{align}
where $w^1_{c_1}$ and $w^2_{c_2}$ are respectively words from the first order and the second order vocabularies.

At the initial stage of search, the inner-products between query and all the vocabulary words are calculated and kept in a table. In addition, for the third term in Eqn.~\ref{eqn:q2key}, it involves the calculation of inner product between $w^1_{c_1}$ and $w^2_{c_2}$. In order to facilitate fast calculation, the inner products between words from the first order vocabulary and the words from the second order vocabulary are pre-calculated and kept for looking up. As a result, fast calculation of Eqn.~\ref{eqn:q2key} is achievable by performing several look-up operations. 

In order to speed-up the search, it is more favorable to calculate the distance between query and the first order codes first. Thereafter, the early pruning is undertaken by ignoring keys whose first order codes are far away from the query. To achieve this, Eqn.~\ref{eqn:q2key} is rewritten as
\begin{align}
d(q, I) = \underbrace{q{\cdot}q^t - 2{\cdot}w^{(1)}_{c_1}{\cdot}q^t - {w^{(1)}_{c_1}}^2}_{term~1} \nonumber \\
	 \underbrace{- 2{\cdot}w^{(2)}_{c_2}{\cdot}q^t-w^1_{c_1}{\cdot}{w^{(2)}_{c_2}}^t-{w^{(2)}_{c_2}}^2}_{term~2}.
\label{eqn:q2key2}
\end{align}
As shown in Eqn.~\ref{eqn:q2key2}, the search process will calculate ``\textit{term 1}'' first. Only keys that are ranked at top are joined into the distance calculation in the 2nd term. Based on $d(q, I)$, the indexing keys are sorted again. Only the inverted lists that are pointed by the top-ranked keys are visited. Note that this cascade pruning scheme is extensible to the case that more than two orders of codes are used as the indexing key. In each inverted list, only the IDs of vectors which share the same RVQ codes are kept. 

\begin{figure}
\begin{center}
   \includegraphics[width=0.95\linewidth]{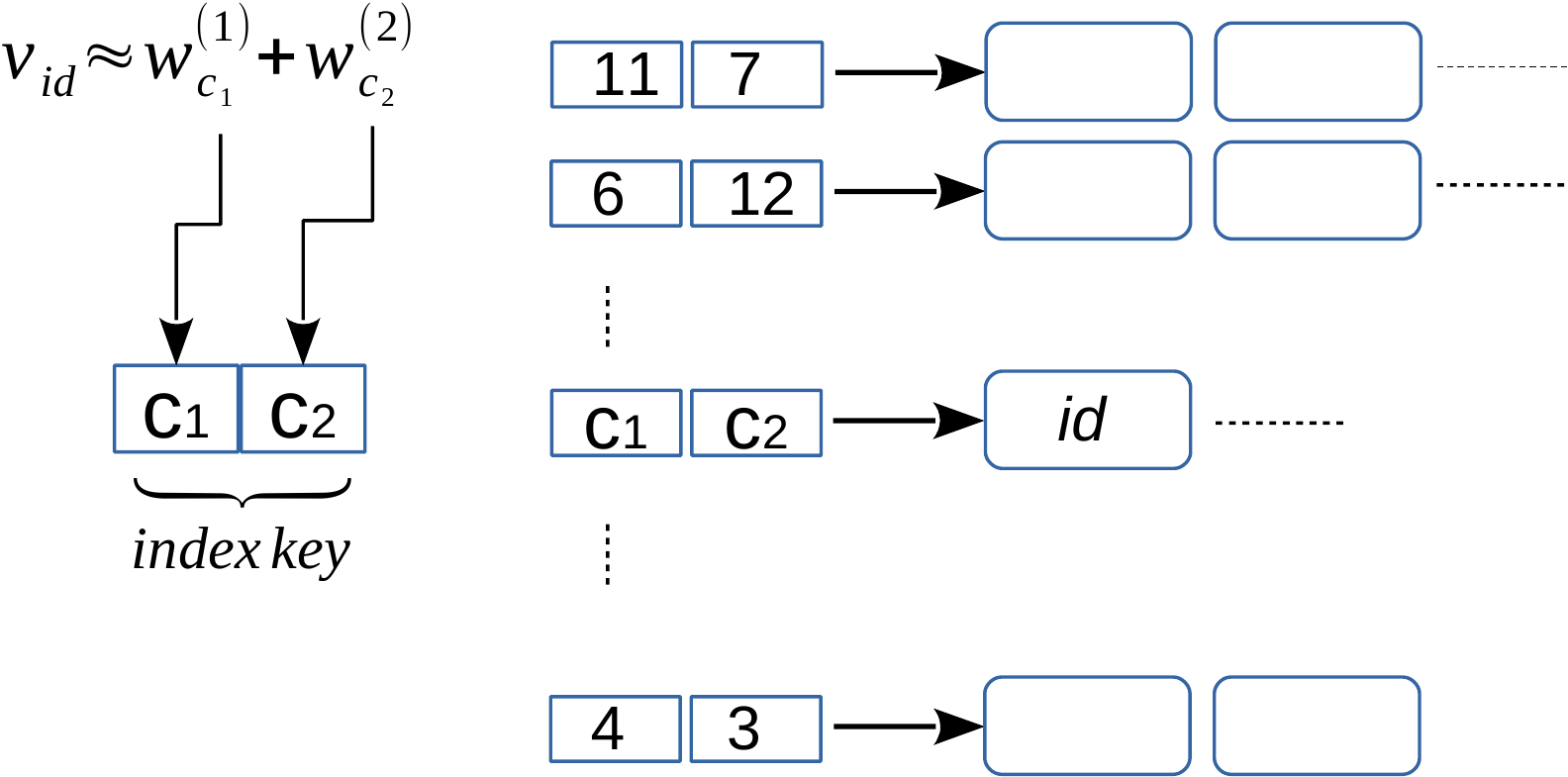}
\caption{An illustration of inverted indexing derived from residue vector quantization. Only two-layer case is demonstrated, while it is extensible to more than two-layer cases. RVQ is shown on the left, the inverted indexing with RVQ codes is shown on the right. Vector $v_{id}$ is encoded by $w^{(1)}_{c_1}$ and $w^{(2)}_{c_2}$, which are the words from the first and the second layer vocabularies respectively.}
\label{fig：egnns}
\end{center}
\end{figure}

In the above analysis, we only show the case that inverted indexing keys produced from the first two orders of RVQ codes. However, it is  extensible to the case of more orders of RVQ codes are employed. It is actually possible to generate key space of trillion level with only four orders of RVQ codes. For instance, if vocabulary sizes of the first four orders are set to $2^{13}$, $2^8$, $2^8$ and $2^8$, the volume of key space is as big as $2^{37}$, which is sufficient to build inverted indices for trillion level reference set. In practice, the number of layers to be employed closely related to the size of reference set. 

Intuitively, the larger of the reference set, the more number of layers should be incorporated. According to our observation, for million level dataset, two-layer RVQ makes a good trade-off between efficiency and search quality. During the online search, all the candidate vectors that reside in the top-ranked inverted lists are collected as the candidate seeds. The seeds are organized as a priority queue according to $d(q, I)$ and are fed to Alg.~\ref{alg:egnns}.

\section{Experiment}
\label{sec:exp}
In this section, the experiments on SIFT1M and GIST1M are presented. In SIFT1M and GIST1M, there are 10,000 and 1,000 queries respectively. In the first two experiments, following the practice in~\cite{efanna16}, the curve of average recall against total search time (of all queries) is presented to show the trade-off that one method could achieve between search quality and search efficiency. The curve is plotted by varying the number of samples to be expanded in each round of top-ranked list and the number of iterations.

We mainly study the performance of the enhanced GNNS (E-GNNS) and the performance of E-GNNS when it is supported by inverted indexing (IVF+E-GNNS). Their performance is presented in comparison to state-of-the-art methods such as original GNNS~\cite{icai11:kiana}, EFANNA~\cite{efanna16}, which is the most effective method in recent studies, FLANN~\cite{pami14:flann}, E2LSH~\cite{lsh04} and ANN~\cite{kdtreesrc}. For E-GNNS and IVF+E-GNNS, we use the same scale of kNN graphs as EFANNA in the experiments, namely 30 and 50 for SIFT1M  and GIST1M respectively. For IVF-E-GNNS, two layers of RVQ quantizer are adopted and their vocabulary sizes are set to 256 for each on both SIFT1M and GIST1M. All the experiments have been pulled out by a single thread on a PC with 3.6GHz CPU and 16G memory setup.

\subsection{E-GNNS versus GNNS}
In our first experiment, we try to show the performance improvement achieved by E-GNNS over GNNS. As shown in Figure~\ref{fig：egnns}, with the revised procedure (Alg.~\ref{alg:egnns}), significant performance boost is observed with E-GNNS. On one hand, E-GNNS takes less number of iterations to converge. On the other hand, it becomes less likely to be trapped in a local optima as it is able to reach to considerably higher recall.

\begin{figure}
\begin{center}
    \subfigure[SIFT1M]
    {\includegraphics[width=0.57\linewidth]{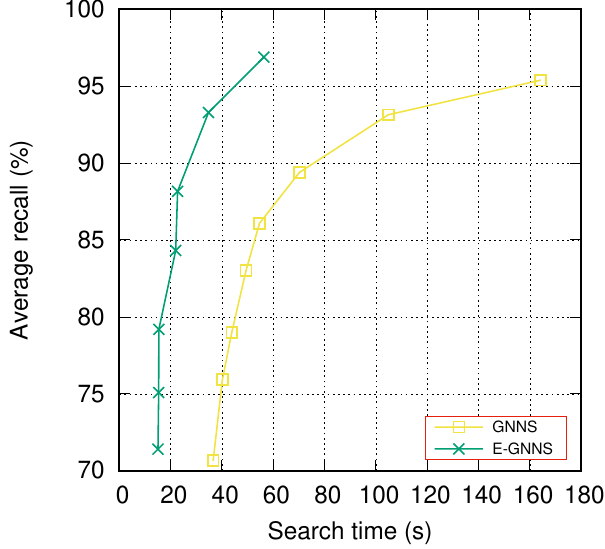}}
    \\
    \subfigure[GIST1M]
    {\includegraphics[width=0.57\linewidth]{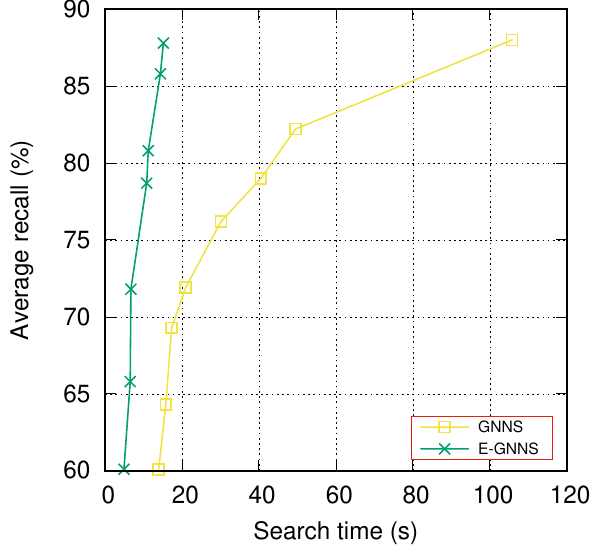}}
\caption{NNS quality against search time for GNNS and E-GNNS. The scale of kNN graph is fixed to 30 and 50 for SIFT1M and GIST1M respectively.}
\label{fig：egnns}
\end{center}
\end{figure}

\subsection{E-GNNS supported by Inverted Index}
\begin{figure}
\begin{center}
    \subfigure[SIFT1M]
    {\includegraphics[width=0.58\linewidth]{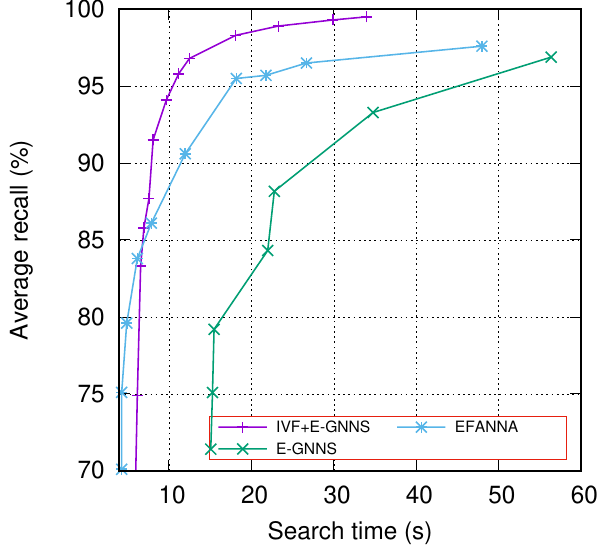}}
    \\
    \subfigure[GIST1M]
    {\includegraphics[width=0.58\linewidth]{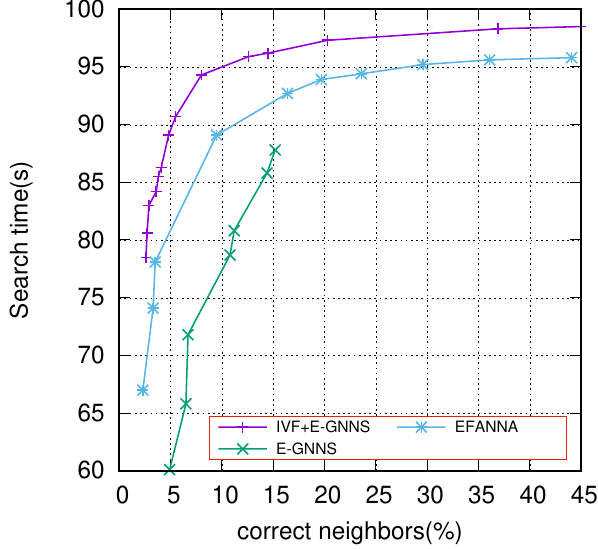}}
\caption{NNS quality against search time for E-GNNS, IVF+E-GNNS and EFANNA. The scale of kNN graphs in SIFT1M and GIST1M is fixed to 30 and 50 respectively for all the methods. }
\label{fig：ivfegnns}
\end{center}
\end{figure}

In the second experiment, we study the performance of E-GNNS when it is supported by inverted indexing derived from RVQ. The performance of EFANNA is also presented for comparison, which similarly follows the scheme of GNNS. Comparing to EFANNA, we adopt inverted indexing  instead of multiple k-d trees to direct the search. Moreover, the enhanced GNNS procedure is adopted in our case. As shown in Figure~\ref{fig：ivfegnns}, E-GNNS supported by inverted indexing shows considerable performance improvement. With similar search time, its average recall is boosted more than 10\% across two datasets. Compared to EFANNA, IVF+E-GNNS converges to significantly higher recall in shorter time. While for EFANNA, it takes shorter time than IVF+E-GNNS only when the requirement of search quality is low. In order to reach to high recall, the search procedure could be unexpectedly elongated. We find that its performance could be even degenerated as one chooses larger expansion factor (comparable to top-k in IVF+E-GNNS). This mainly indicates GNNS directed by multiple k-d trees is easier to be trapped in local optima than that of IVF+E-GNNS. Notice that IVF+E-GNNS also outperforms the method proposed in~\cite{mm12:jdwang} by a large margin as it is reported in~\cite{efanna16}.

\subsection{Comparison to Non-compressional Methods}
\begin{table}
\caption{Parameter settings for non-compressional methods}
\scriptsize{
\begin{center}
\begin{tabular}{| c | c |c|}
\hline
& SIFT1M & GIST1M \\
\hline
ANN~\cite{kdtreesrc} & $\varepsilon = 0$ & $\varepsilon = 5 $\\
\hline
E2LSH~\cite{lsh04} & $r = 0.45$ & $r = 1$ \\
\hline
\multirow{2}{*}{FLANN~\cite{pami14:flann}} & p = 0.9,  & p = 0.8, \\
 & $\omega_m = 0.01, \omega_t = 0$ & $\omega_m = 0.01, \omega_t = 0$\\
\hline
EFANNA~\cite{efanna16} & nTree = 4, kNN=30 & nTree = 4, kNN=50\\
\hline

\end{tabular}
\end{center}
}
\label{tab:paraconfig}
\vspace{-0.15in}
\end{table}

\begin{table*}
\caption{Comparison on average time cost per query (ms), memory consumption (relative ratio) and quality (recall@top-k) of all representative NNS Methods in the literature. In terms of memory consumption, the ratio is taken between real memory consumption of one method against the original size of the candidate dataset}
\scriptsize{
\begin{center}
\begin{tabular}{|l|c|c|c|c||c|c|c|c|} \hline
&  \multicolumn{4}{c|}{SIFT1M} & \multicolumn{4}{c|}{GIST1M} \\\hline
&  T (ms) & Memory &R@1& R@100 & T (ms) & Memory & R@1& R@100 \\\hline\hline
IVFPQ-8~\cite{JDS11} & 8.6 &0.086{$\times$} & 0.288 & 0.943 & 36.2 & 0.0029{$\times$} & 0.082 & 0.510 \\ \hline
IVFRVQ-8~\cite{ivfrvq10} & 4.7 & 0.070{$\times$} & 0.270 & 0.931 & 12.5 & 0.0023{$\times$} & 0.107 & 0.616 \\ \hline
IMI-8~\cite{pami15:artem} & 95.1 & 0.086{$\times$} & 0.228 & 0.924 & 90.0 & 0.0029{$\times$} & 0.053 & 0.369\\ \hline \hline
FLANN~\cite{pami14:flann} & 0.9 & 6.25{$\times$} & 0.852 & 0.852 & 11.9 & 2.12{$\times$} & 0.799 & 0.799\\ \hline
ANN~\cite{kdtreesrc} &  250.5 & 8.45{$\times$} & 0.993 & 1.000 & 232.0 & 1.48{$\times$} & 0.938 & 1.000\\ \hline
E2LSH~\cite{lsh04} & 30.0 & 29.10{$\times$} & 0.764 & 0.764 & 6907.7 & 1.74{$\times$} & 0.342 & 0.990\\ \hline 
E-GNNS~ & 2.3 & 1.225{$\times$} & 0.882 & 0.882 & 6.7 & 1.049{$\times$} & 0.724 & 0.724\\ \hline
IVF+E-GNNS~ & 1.8  & 1.234{$\times$} & 0.983 & 0.983 & 8.0 & 1.053{$\times$}  & 0.943 & 0.943 \\ \hline
EFANNA~\cite{efanna16}~ & 1.8 & 1.621{$\times$} & 0.955 & 0.955 & 9.5 & 1.049{$\times$} & 0.891 & 0.891\\ \hline \hline

Exhaustive NNS & 1,107.0 & 1.0{$\times$} & 1.00 & 1.00 & 2,258.0 & 1.0{$\times$} & 1.00 & 1.00 \\ \hline
\end{tabular}
\end{center}
}
\label{tab:allnns}
\vspace{-0.1in}
\end{table*}

In this experiment, the performance of IVF+E-GNNS is compared to both compressional methods such as IVFPQ~\cite{JDS11}, IVFRVQ~\cite{ivfrvq10} and IMI~\cite{pami15:artem} and non-compressional methods such as E2LSH~\cite{lsh04}, ANN~\cite{kdtreesrc} and FLANN~\cite{pami14:flann}. The speed efficiency, search quality and memory cost of all these methods on SIFT1M and GIST1M are considered and shown in Table~\ref{tab:allnns}. In particular, the memory cost is shown as the ratio between the memory consumption of one method and the size of reference set. All results reported for the non-compressional methods are obtained by running the codes provided by the authors. The configuration on one method is set to where balance is made between speed efficiency and search quality (see detailed configurations in Table~\ref{tab:paraconfig}). The performance from exhaustive NNS is supplied for reference.

As seen from Table~\ref{tab:allnns}, ANN shows stable search quality and best trade-off on two datasets in terms of its recall, however its speed is only several times faster than brute-force search. Among all methods considered in this experiment, IVF+E-GNNS shows the best trade-off between memory cost, speed efficiency and search quality. Additionally, comparing with all other non-compressional methods, its memory consumption is only slightly larger than the size of reference set. 

\section{Conclusion}
We have presented our solution for efficient nearest neighbor search, which is based on hill-climbing strategy. Two major issues have been discussed in the paper. Firstly, a novel kNN graph construction method is proposed. Its time complexity is only at $O(n{\cdot}log(n){\cdot}D)$, which makes it scalable to very large scale dataset. With the support of constructed kNN graph, an enhanced hill-climbing procedure has been presented. Furthermore, with the support of inverted indexing, it shows superior performance over most of the NNS methods in the literature. We also show that this method is easily scalable to billion level dataset with the support of the inverted indexing derived from multi-layer residue vector quantization.

%\section*{Acknowledgement}
%This work is supported by National Natural Science Foundation of China under grants 61572408. The authors would like to express their %sincere thanks to ODD Concepts Inc. from Seoul, South Korea for their generous support.

\bibliographystyle{ieeetr}
\bibliography{wlzhao}

\end{document}